\documentclass[letterpaper, 10 pt, conference]{ieeeconf}
\IEEEoverridecommandlockouts                              
\usepackage{comment}

\usepackage{algorithm}
\usepackage{algorithmic}
\usepackage{url}
\usepackage{amsmath}
\usepackage{cite}
\usepackage[dvips]{graphicx}
\usepackage{epsfig}
\usepackage{epic,eepic,color}
\usepackage{times}
\usepackage{mathptmx}
\usepackage{multirow}
\usepackage{stkernel}

\usepackage{cases}

\usepackage{times}
\usepackage{multirow}

\definecolor{red}{rgb}{1,0,0}
\definecolor{green}{rgb}{0,1,0}
\definecolor{blue}{rgb}{0,0,1}
\definecolor{violet}{rgb}{1,0,1}
\definecolor{cyan}{cmyk}{1,0,0,0}
\definecolor{magenta}{cmyk}{0,1,0,0}
\definecolor{yellow}{cmyk}{0,0,1,0}

\definecolor{white}{rgb}{1,1,1}

\usepackage{jumoline}

\newcommand{\CO}[1]{}

\newcommand{\CommentOut}[1]{}

 \newcommand{\editage}[1]{}

\newcommand{\FIG}[3]{
\begin{minipage}[b]{#1cm}
\begin{center}
\includegraphics[width=#1cm]{#2}\\
{\scriptsize #3}
\end{center}
\end{minipage}
}

\newcommand{\FIGU}[3]{
\begin{minipage}[b]{#1cm}
\begin{center}
\includegraphics[width=#1cm,angle=180]{#2}\\
{\scriptsize #3}
\end{center}
\end{minipage}
}

\newcommand{\FIGR}[3]{
\begin{minipage}[b]{#1cm}
\begin{center}
\includegraphics[angle=-90,clip,width=#1cm]{#2}
\\
{\scriptsize #3}
\vspace*{1mm}
\end{center}
\end{minipage}
}

\newcommand{\FIGpng}[5]{
\begin{minipage}[b]{#1cm}
\begin{center}
\includegraphics[bb=0 0 #4 #5, clip, width=#1cm]{#2}\vspace*{-1mm}\\
{\scriptsize #3}
\vspace*{1mm}
\end{center}
\end{minipage}
}

\renewcommand{\FIGpng}[5]{~} \renewcommand{\FIGU}[3]{~} 

\newcommand{\figA}{
\begin{figure}[t]
  \begin{center}
\FIG{8}{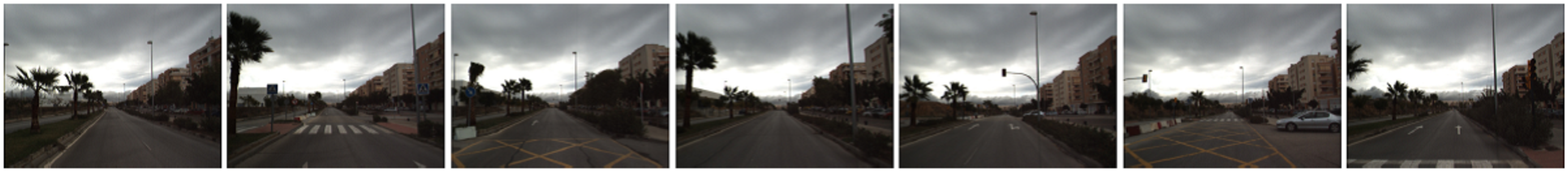}{}
\caption{Examples of places selected by UPD.}\label{fig:A}
\end{center}
    \end{figure}
}

\newcommand{\figB}{
\begin{figure}[t]
  \begin{center}
\FIG{8}{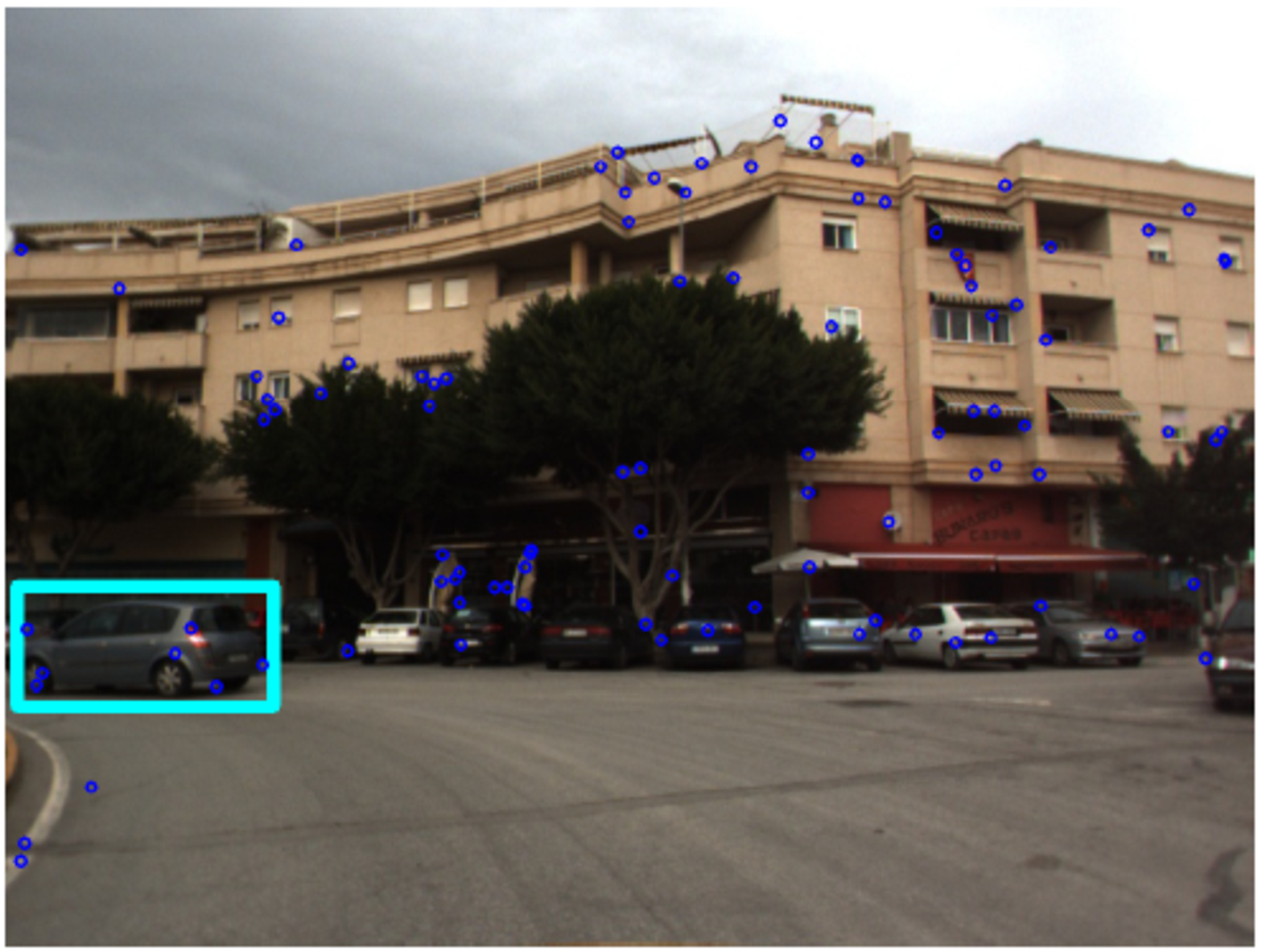}{}
\caption{Single-view change detection.}\label{fig:B}
\vspace*{-3mm}
  \end{center}
\end{figure}
}

\newcommand{\figC}{
\begin{figure}[t]
  \begin{center}
\FIG{8}{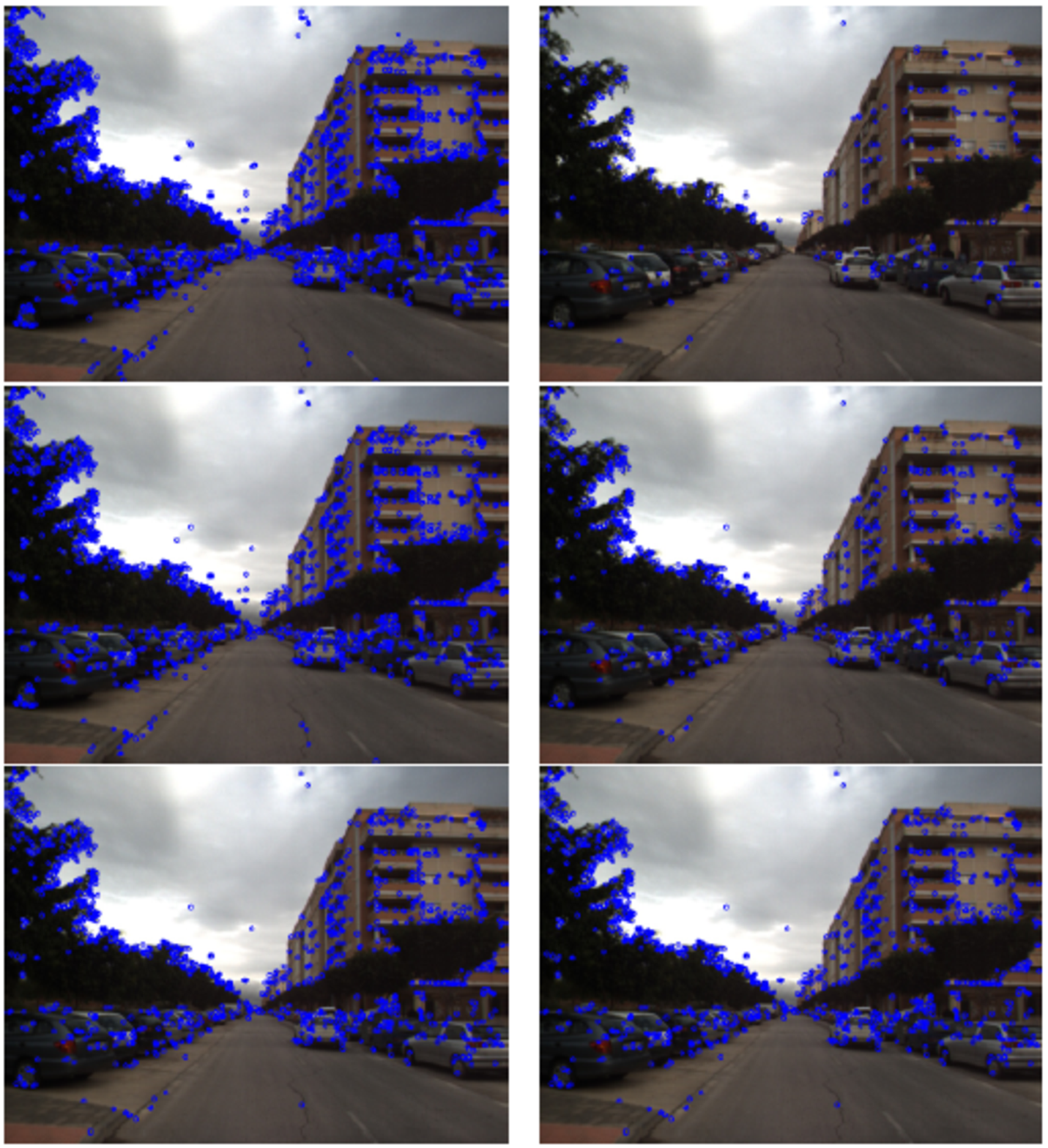}{}
\caption{Examples of nuisance prediction.}\label{fig:C}
  \end{center}
\end{figure}
}

\newcommand{\figD}{
\begin{figure}[t]
  \begin{center}
\FIG{8}{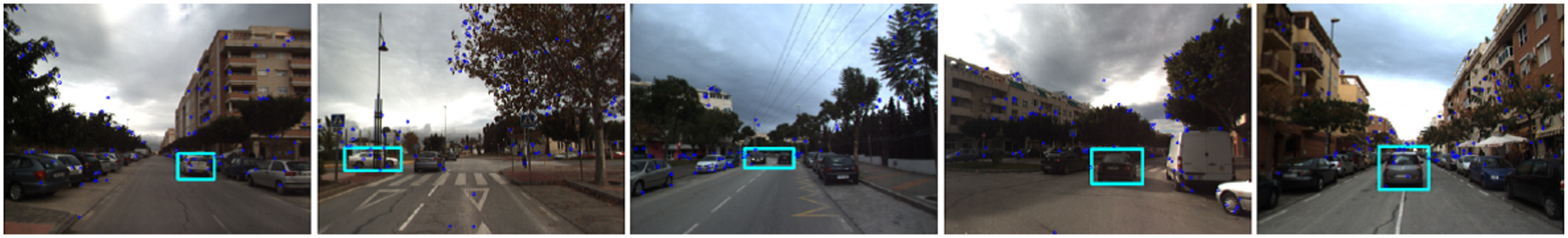}{}
\caption{Examples of change detection.}\label{fig:D}
  \end{center}
\end{figure}
}

\newcommand{\figE}{
\begin{figure}[t]
  \begin{center}
\FIGR{8}{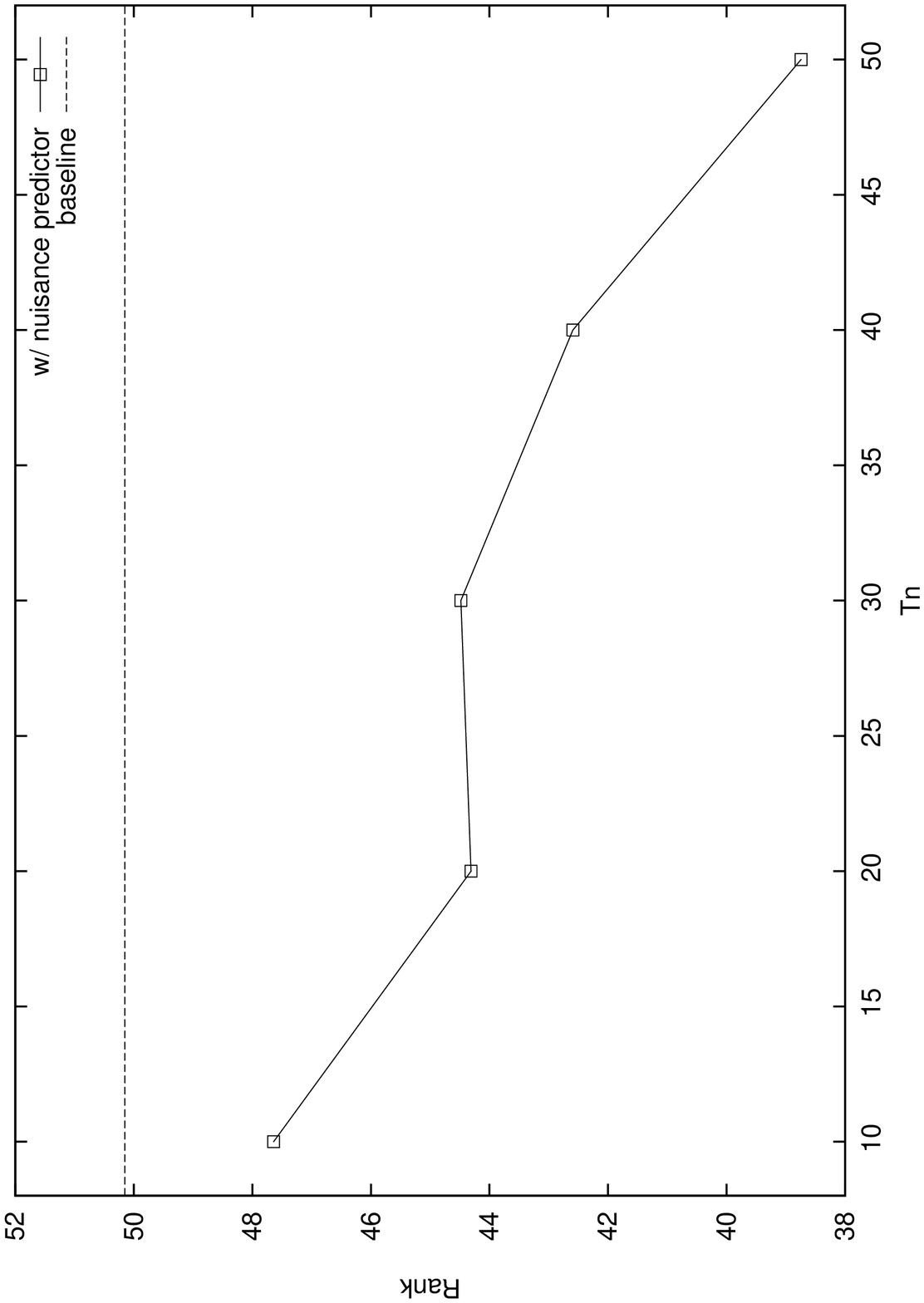}{a}
\FIG{8}{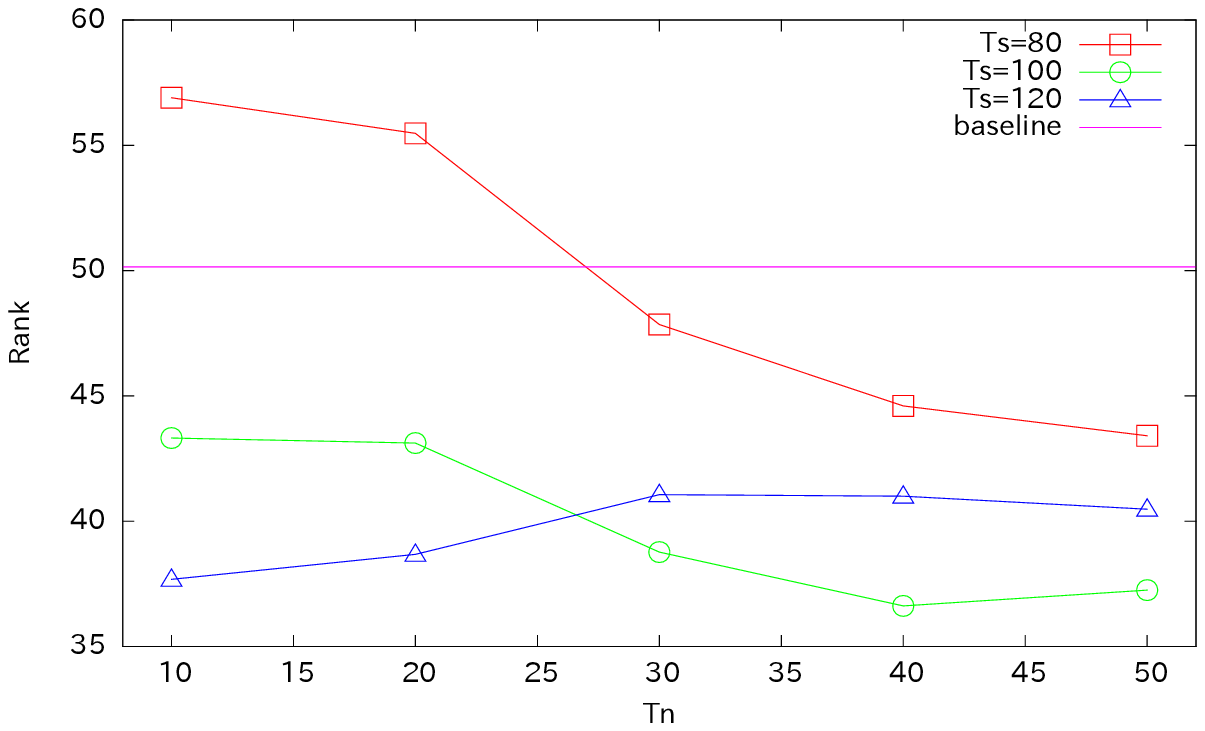}{b}
\caption{Quantitative results.}\label{fig:E}
  \end{center}
\end{figure}
}

\newcommand{\figF}{
\begin{figure}[t]
  \begin{center}
\FIG{8}{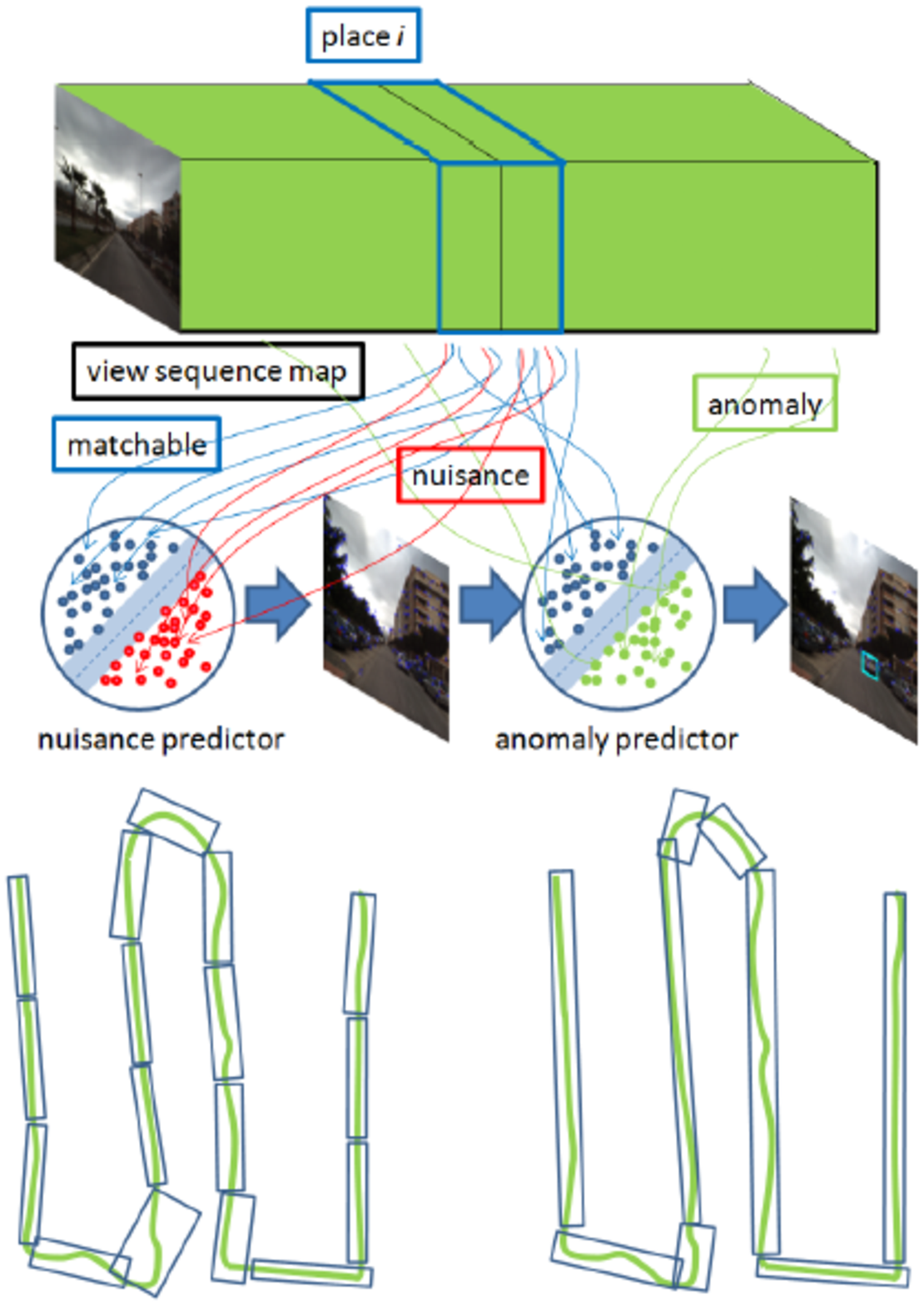}{}
\caption{Supervised change detection using place-specific change classifiers and unsupervised place discovery (UPD). Top: change detection pipeline consisting of nuisance and anomaly predictors. Bottom: two strategies for UPD: time cue strategy (left) and appearance cue strategy (right). The curved line and bounding boxes indicate the robot's trajectory and individual place regions determined by either UPD strategy.}\label{fig:F}
  \end{center}
\end{figure}
}

\newcommand{\tabA}{
  \begin{table}[t]
    \begin{center}
      \caption{Performance results}\label{tab:A}
      \vspace*{2mm}
  \begin{tabular}{l|ll}
\hline
Method & Rank & \\
\hline
\hline
baseline & 50.15 & ($\pm$ 0.00) \\
\hline
random ref & 63.85 & (- 13.70) \\
\hline
Tn=30 Ts=100 & 38.77 & (+ 11.38) \\
\hline
Tn=30 & 44.51 & (+ 5.64) \\
\hline
Ts=80 & 51.43 & (- 1.283) \\
(time cue) & 61.38 & (- 11.23) \\
\hline
Ts=100 & 46.61 & (+ 3.54) \\
(time cue) & 54.97 & (- 4.82) \\
\hline
Ts=120 & 48.06 & (+ 2.09) \\
(time cue) & 48.78 & (+ 1.37) \\
\hline
  \end{tabular}
    \end{center}
  \end{table}
}

\newcommand{\tabB}{
  \begin{table}[t]
    \caption{Number of places}\label{tab:B}
      \vspace*{2mm}
    \begin{center}
  \begin{tabular}{l|l}
\hline
Ts & \# SVMs \\
\hline
\hline
$\infty$ &  1731 \\
120 &  947 \\
100 &  657 \\
80 &  413 \\
\hline
  \end{tabular}
    \end{center}
  \end{table}
  }

\begin{document}

\title{
Unsupervised Place Discovery for Place-Specific Change Classifier
}

\author{Fei Xiaoxiao ~~~~~~ Tanaka Kanji
\thanks{Our work has been supported in part by 
JSPS KAKENHI 
Grant-in-Aid for Young Scientists (B) 23700229,
and for Scientific Research (C) 26330297 
}
\thanks{The authors are with Graduate School of Engineering, University of Fukui, Japan.
{\tt\small tnkknj@u-fukui.ac.jp}}%
\vspace*{-5mm}}

\maketitle

\begin{abstract}
In this study, we address the problem of supervised change detection for robotic map learning applications, in which the aim is to train a place-specific change classifier (e.g., support vector machine (SVM)) to predict changes from a robot's view image. An open question is the manner in which to partition a robot's workspace into places (e.g., SVMs) to maximize the overall performance of change classifiers. This is a chicken-or-egg problem: if we have a well-trained change classifier, partitioning the robot's workspace into places is rather easy. However, training a change classifier requires a set of place-specific training data. In this study, we address this novel problem, which we term unsupervised place discovery. In addition, we present a solution powered by convolutional-feature-based visual place recognition, and validate our approach by applying it to two place-specific change classifiers, namely, nuisance and anomaly predictors.
\end{abstract}

\figF

\section{Introduction}

In this study, we address the problem of supervised change detection in the domain of robotic map learning applications \cite{cadena2016past}, in which the aim is to train a place-specific change classifier (e.g., support vector machine (SVM)) to predict changes from a robot's view image. Change detection is crucial to the success of long-term map learning and updating in realistic dynamic environments. Specifically, we want to design an approach that can scale to large maps and that can function under incremental additions and deletions of individual places. Addressing this problem at large scale is critical, particularly in the context of long-term map learning, because of the requirements of incremental map learning.

One open question is the manner in which to partition a robot's workspace into places (e.g., SVMs) in order to maximize the overall performance (e.g., accuracy, precision, recall) of change classifiers (Fig. \ref{fig:F}). We call this the problem of unsupervised place discovery (UPD). UPD is a critical issue, as the definition of places strongly influences the performance of a change classification task. Intuitively, each place should be defined as a continuous region in the robot's workspace having similar visual appearance. UPD is a chicken-or-egg problem: if we have a set of well-trained place-specific classifiers, partitioning the robot's workspace into place regions is rather easy. However, the training of a change predictor requires a set of pre-defined place classes.

To address this issue, we develop a UVP technique powered by deep convolutional neural network (DCN)-based visual place recognition (VPR) \cite{lowry2016visual}. Our approach is motivated by the recent success of DCN in various scene-recognition tasks \cite{alexnet}. In \cite{babenko2014neural}, the output of an intermediate fully connected layer of DCN is used as a semantic image descriptor and further compressed to realize efficient image retrieval. In \cite{paulin2015local}, the output of a convolutional layer of DCN is viewed as an array of local feature descriptors and further encoded to realize the bag-of-words model. We propose using the local feature approach, because it is more suited than the global image descriptor approach to recognizing changeable environments where the appearance may be locally different in the training and test data.

We formulate change detection as a supervised recognition task. In general, change detection techniques are classified into two broad categories: unsupervised \cite{bruzzone2000automatic} and supervised \cite{nielsen2007regularized}. Unsupervised techniques aim to differentiate two images in order to detect changes that have occurred between two dates, and hence are suited to general purpose change detection for applications such as satellite imagery \cite{bovolo2008novel}. Supervised techniques aim to learn and predict a given labeled training set, and hence are suited to application-aware change detection such as a 3D cadastral city model \cite{taneja2015geometric}. We observe that supervised approaches are more suited to our application domain of long-term map learning, in which we treat the available 
map as a rich information source for mining labeled training data for both normal features (i.e., non-changes) and anomalies (i.e., changes).

As an additional contribution, we present a practical system for supervised change detection. In general, what constitutes interesting and uninteresting changes (i.e., nuisances) depends on place-specific contextual information. We focus on two types of place-specific classifiers: nuisance and anomaly predictors. A nuisance predictor aims to learn the characteristics of non-nuisance features (i.e., change candidates) by detecting, tracking, and learning a set of matchable features (i.e., non-nuisances) over an available map. An anomaly predictor aims to learn the characteristics of anomalies (i.e., changes) from normal visual features and anomalies that are mined from a map. Note that our approach requires the entire map only at the offline training stage and does not require it as query input nor as part of the classifier's data structure. 

We conduct experimental evaluation on the publicly available Malaga dataset \cite{blanco2014malaga} (Fig. \ref{fig:B}). Our experiments show that our algorithm can detect both small and large changes in input scenes. In addition, our UPD method is found to be effective at improving the accuracy of place-specific change detection while maintaining the number and compactness of the place-specific predictors. In addition to the aforementioned contributions, our experimental system can also be viewed as a novel solution to the moving object detection (MOD) alternative task, which complements existing MOD approaches based on motion cues (e.g., motion segmentation, moving camera background subtraction) and appearance cues (e.g., particular moving object recognition).

\figB

\section{Relation to Existing Works}

This study is related to existing studies on change detection \cite{radke2005image}. In \cite{bruzzone2000automatic}, two types of Bayesian approaches were presented that minimize overall change detection error probability under the independent pixel assumption and exploit interpixel class dependency contexts. In \cite{alcantarilla2016streetview}, a deep deconvolutional network method for pixel-wise change detection was trained. However, the aforementioned methods are based on the assumption that the pair of input and reference images are coarsely registered prior to the change detection. In our application domain of robotic map learning applications, the registration of partially overlapping images under dynamic scenes is itself difficult and thus represents an open problem \cite{lowry2016visual}. In \cite{taneja2015geometric}, this issue of partial view overlap was addressed by considering the individual alignment error of an image with a 3D model, as well as the relative alignment error of an image in relation to its neighbors. However, this method requires that a 3D cadastral city model be given, and obtaining such a 3D model is not a trivial task in the case of autonomous robotics applications. In \cite{kovsecka2012detecting}, the 3D locations of sparse visual features were estimated from SLAM or SfM techniques and then used as cues for change detection. This method is similar to our own approach in its objectives. However, the approach in 
\cite{kovsecka2012detecting} directly compares the images of the image pair and does not train an efficient place-specific change predictor. Moreover, the problem of the robot's workspace partitioning (i.e., UPD) has not been adequately addressed in previous studies.

\section{Approach}

Although our approach is general and may be applicable to various types of maps including view graphs \cite{konolige2010view}  and 3D point clouds \cite{engelhard2011real}, we focus on simple view sequence maps \cite{matsumoto1996visual}. A view sequence map is one of the most standard map formats widely used in robotic map learning applications \cite{cadena2016past}.

To train place-specific change and nuisance predictors, we propose using support vector machine (SVM) \cite{tong2001support}. One desirable property of SVM is the compactness of the model and efficiency in training the classifier. This property enables the addition of new place-specific predictors during incremental map learning tasks, and also allows for modifying or deleting existing place-specific predictors when the robot identifies inconsistencies between the environment and map (e.g., changes). Of course, we could replace the SVMs with a more powerful prediction model such as a DCN. However, not all DCNs are efficient to train and compact sufficiently in order to serve as place-specific prediction models. To implement the SVM, we adopt SVM light with a linear kernel.

\subsection{Anomaly Predictor}

The anomaly predictor aims to learn no-change features in order to predict changes (i.e., anomalies) in a direct manner. We follow the pipeline of first extracting local feature descriptors (e.g., SIFT) from mapped images, after which these features are fed to an SVM learner. Once the training stage is finished, we use the anomaly predictor to rank input features in descending order of the SVM's output. Top-ranked features are then regarded as change, which represent the final output of our change classification system.

\figC

Note that we use SIFT as the local feature for anomaly prediction. In our preliminary experiment, we also tested local features from DCN, local convolutional features (LCF), and we found that LCFs often fail to detect small change regions because of the low spatial resolution (e.g., 32 pixels) of those LCFs. We reserve the following for future work: to explore additional powerful DCN-based features and to exploit the change predictor for our own method's advantage.

\figA

\subsection{Nuisance Predictor}\label{sec:np}

The nuisance predictor is trained for single-view nuisance detection. It aims to filter out false positive changes prior to the anomaly prediction stage. Training examples are harvested during the offline training stage by applying a local descriptor matcher to each successive frame pair in the view sequence map, and then matched features are used as examples of normal (i.e., non-nuisance) features. For the local descriptor matcher, we use a nearest neighbor matcher with L2 distance and a consistency check, in which the retrieved feature is checked if the query feature is also the one-nearest neighbor feature for the retrieved feature. Once the training stage is finished, we use the nuisance predictor to rank input features in descending order of the probability of nuisance, and then the top-$T_n$ [\%] ranked features are regarded as nuisance to be filtered out.

Note that the nuisance predictor deals with a one-sided classification problem, in which only negative examples (i.e., non- nuisance) are given during the training stage. Unlike typical semi-supervised settings such as active learning, no labeled data are available for the positive class. To address this issue, we consider the task of nuisance mining, the aim of which is to mine nuisance features from a large collection of visual features. For the feature collection, we introduce a large image set that is independent from query and mapped images and which we term ``visual experience" (see Section \ref{sec:exp} for details).

Our mining strategy is to search for those samples that are most distant from each negative feature (i.e., non-nuisances) in $S$. This means that distant samples should be positive with high probability. More formally, each pseudo positive sample is mined by means of the following
\begin{equation}
s^* = arg \max_{s\in S} | q - s|_2,
\end{equation}
from each negative instance $q$ in the query image. Here, $|\cdot|_2$ is L2 norm and $S$ is the feature set from the visual experience.

Fig. \ref{fig:C} visualizes the output of the nuisance predictor. The top, middle, and bottom panels correspond to cases in which the nuisance threshold percentage $T_n$ is set to 10, 30, and 50, respectively. For each case, the left and right figures show keypoints of features classified as non-nuisance and nuisance, respectively. It can be seen that our nuisance predictor successfully classified items such as plants or building edges into the nuisance class, and that a non-negligible number of false positive/negative nuisances exist. Note that visual features representing nuisance features are place-dependent and our classifier is successful at classify these features in the examples shown.

\section{UPD}

We developed two basic strategies for UPD. Fig. \ref{fig:A} shows representative images for a length-7 sub-sequence of a place sequence that is selected by the UPD algorithm described in Subsection \ref{sec:acs} that follows.

\subsection{Time Cue Strategy}\label{sec:tcs}

The first strategy is a simple time cue strategy, which partitions a sequence of images into places based on their time stamps or image IDs. We uniformly sample $K-1$ partitions from the space of image IDs and obtain a set of $K$ clusters with approximately equal intervals. This strategy is based on the observation that images having similar time stamps are expected to be similar in appearance. This is because they are collected by a mapper robot that navigates through a continuous trajectory in the environment. In addition, such a cluster of images is expected to be a good training set for a place-specific classifier. Obviously, this simple strategy has many limitations. Specifically, it is not robust against variations in a robot's moving speed. More importantly, it does not take advantage of any appearance cues that are available from the map.

\subsection{Appearance Cue Strategy}\label{sec:acs}


The second strategy is the appearance cue strategy. Essentially, this strategy augments the time cue strategy by using the available appearance cue from the map. It begins by initializing a single keyframe with the first image frame. It evaluates the similarity of visual features between each image frame and the latest keyframe, and if the image-level dissimilarity is large, we consider the two frames belong to different place regions. We then start the next place region by initializing a new keyframe. To check the dissimilarity between two frames, we extract a bag of LCFs from either frame, and then employ nearest neighbor matching to obtain feature correspondences. For the LCFs, we use a 31 $\times$ 23 array of 512-dim feature vectors from the last convolutional layer of the DCN as the image representation, as it showed excellent performance in the image classification task in \cite{paulin2015local}. The image-level dissimilarity value is then evaluated by an image-to-class distance metric and compared against a pre-defined threshold $T_s$. If it is lower than the threshold, the image pair is considered to belong to the same place. For the image-to-class distance metric, we base it on the naive Bayes nearest neighbor to compute the distance from the query image to the place class (represented by the keyframe) using L1 norm to measure the feature-level distance.

\figD

\figE

\section{Experiments}\label{sec:exp}

We evaluated the performance of place-specific change classification and UPD in highly dynamic urban environments. 
For our evaluation, we used image sequences from the Malaga dataset \cite{blanco2014malaga}. The Malaga dataset contains GPS data, image sequences from a stereo camera facing forward in the direction of a robot car's motion, as well as LIDAR data. We used the left camera's images having a resolution of 1024 $\times$ 768 for our change classification task. To guarantee that at least one relevant mapped image exists for every query image, we considered a practical place recognition scenario called loop closing \cite{cadena2016past}, which was borrowed from the field of visual SLAM, in which a robot traverses a loop-like trajectory and then returns to the previously explored location. More formally, the relevant image pair was defined by an image pair that satisfies two conditions: 1) the viewpoint from each query is nearer than that of other candidates. 2) the distance traveled along the robot's trajectory is sufficiently greater (400 m) than that of the query image. Because of Condition 2, every relevant image pair becomes a dissimilar image pair, which represents a challenge for our UPD and change classification tasks. In our experiments, we used datasets \#s 5-8, and \#10 from the Malaga dataset because they contain potential query images that satisfy the aforementioned loop-closing conditions. These datasets contain 4816, 4618, 2121, 10026, and 17310 images, respectively. For the visual experience described in Subsection \ref{sec:np}, we used Malaga dataset \#5/\#6/\#7/\#8/\#10 for the visual experience of dataset \#10/\#5/\#6/\#7/\#8.

In the first experiment, the effectiveness of UPD was investigated. The performance of the change classification algorithm was evaluated using a set of query images. For the query images, we used 109 random images that satisfied the aforementioned two loop-closing conditions such that for each moved object (e.g., the other car) that the camera encountered, one or a few query images appeared. The output of the change classification algorithm was a collection comprising the likelihood of change (i.e., SVM's output) for every local feature in every query image. The features in a query image were sorted in descending order of likelihood of change. The rank values of features that belong to the ground-truth changed objects with respect to the sorted feature list were then used as a measure for performance evaluation. For the ground truth, we manually annotated changed objects in the form of bounding boxes (Fig. \ref{fig:B}) by comparing query and reference images. Because the evaluation was based on ranking, a smaller value signified better performance. If multiple local features belonged to the ground-truth bounding box, the rank value of the feature that was assigned the largest likelihood of change was used for the evaluation.

Fig. \ref{fig:D} shows example results of change classification. The ground truth changed objects and top-100 ranked change classification results are indicated by bounding box and points. The proposed method was successful in distinguishing changed from non-changed objects in these examples. However, a major source of false positives in the change classification task derived from the fact that edges from plants or buildings were often classified as non-nuisance and thus independently considered as candidate changes in the next anomaly prediction stage. In particular, the recognition system had difficulty distinguishing moving cars from those parked, a task that is rather easy for a human vision system. One reason for the false positives is the lack of semantic reasoning as well as place-specific deep reasoning, which are directions for future work.

\tabA
\tabB

Fig. \ref{fig:E} shows quantitative results. We compared several different techniques and pipelines for change classification. 
The ``Rank" indicates the rank value averaged over all the 109 qeury images.
The baseline method (``baseline") was a simple approach that used neither the nuisance predictor nor the UPD method. Thus, it used a non-compact prediction model that consists of a large number of anomaly predictors. Here, an anomaly predictor was trained and stored after every 10 frames in the view sequence map.

Fig. \ref{fig:E}a shows change classification performance based on different settings of the nuisance predictor's parameter $T_n$[\%] of 10, 20, 30, 40, 50 as well as for those cases involving the baseline method (horizontal line). Overall, we observed that the performance was better when the nuisance predictor was used than when not used. In particular, the nuisance predictor with $T_n$ = 50 [\%] outperformed the baseline by a wide margin of 11 points. Note that the nuisance predictor is far from perfect (as shown in Fig. \ref{fig:C}) and thus a non-negligible number of false negative matches occurred. Despite the imperfection, the nuisance predictor was effective at improving change classification performance.

Fig. \ref{fig:E}b shows change classification performance when the nuisance predictor and UPD strategies are combined. We can see that the nuisance predictor was effective when the parameter $T_n$ was sufficiently large (e.g., $T_n$ $>$ 30[\%]). The setting $T_s$ = 80 with a small $T_n$ value was worse than in the baseline method. However, it is noteworthy that compactness of the system considerably improved as the number of place regions (i.e., SVMs) were reduced from 1731 to 413 (See Table \ref{tab:B}).

Table \ref{tab:A}, \ref{tab:B} shows the performance and cost for different methods based on their default parameter settings. The ``random ref" method simulated a simple scenario in which GPS prior was not available, which is relevant in the application of autonomous robotic map learning. Thus, in this method, one place-specific classifier was randomly selected per query and used during the test stage. The method ``$T_n=X$" is used when the nuisance predictor is employed with the parameter $T_n=X$. The method ``$T_s=Y$" is used when places are selected by the appearance cue strategy with the parameter  
$T_s=Y$, whereas ``(time cue)" indicates the case when the same number of places are selected by the time cue strategy. The method ``$T_n=X$ $T_s=Y$" is used when the aforementioned two methods, ``$T_n=X$" and ``$T_s=Y$," are combined. The performance was clearly not good with the ``random ref" case. As can be seen, the results were even worse than the other baseline case when places were densely sampled (i.e., 0.1 [places per frame]). This means that proper selection of place-specific training data and classifiers is critical to successful change classification. We can see that the appearance cue strategy was clearly more effective than the time cue strategy. We also observed that the nuisance predictor (with the parameter $T_n$ = 30) performed better than the appearance cue strategy with the parameters $T_s$= 80, 100, 120). The case of the appearance cue with $T_n=30$ $T_s=100$ was better than that of the time cue with $T_s=30$. This performance was one of the best compared to the other methods considered in our experiment.

\section{Conclusions}

In this study, we addressed the problem of supervised change detection for robotic map learning applications and presented a solution that combines place-specific change classification and UPD. Experimental results reveal that the proposed approach effectively detected changes in highly dynamic scenes and that UPD improved overall change classification performance. It is noteworthy that our experimental system can be viewed as a novel approach for moving object detection (MOD) and that is orthogonal to other existing MOD (e.g., motion-, appearance-, location-based) approaches. Combining these orthogonal approaches to improve overall detection performance is another direction for future research.

\bibliographystyle{IEEEtran}
\bibliography{paper}

\end{document}